\documentclass[runningheads]{llncs}
\usepackage[T1]{fontenc}
\usepackage{graphicx,verbatim}
\usepackage{multirow}
\usepackage{amsmath}
\usepackage{amssymb}
\usepackage{booktabs} 
\usepackage{caption}
\usepackage{pifont}
\usepackage{xcolor}
\begin{document}
\title{Learning to Trim: End-to-End Causal Graph Pruning with Dynamic Anatomical Feature Banks for Medical VQA}
%

\author{Zibo Xu\inst{1} \and Qiang Li\inst{1} \and Weizhi Nie\thanks{Corresponding author.}\inst{2} \and Yuting Su\inst{2}}

\authorrunning{Z. Xu et al.}
\titlerunning{Learning to Trim for Medical VQA}

\institute{School of Microelectronics, Tianjin University, Tianjin 300072, China 
 \and
School of Electrical and Information Engineering, Tianjin University, Tianjin 300072, China\\
\email{weizhinie@tju.edu.cn}}

\maketitle           
\begin{abstract}

Medical Visual Question Answering (MedVQA) models often exhibit limited generalization due to reliance on dataset-specific correlations, such as recurring anatomical patterns or question-type regularities, rather than genuine diagnostic evidence. Existing causal approaches are typically implemented as static adjustments or post-hoc corrections. To address this issue, we propose a Learnable Causal Trimming (LCT) framework that integrates causal pruning into end-to-end optimization. We introduce a Dynamic Anatomical Feature Bank (DAFB), updated via a momentum mechanism, to capture global prototypes of frequent anatomical and linguistic patterns, serving as an approximation of dataset-level regularities.
We further design a differentiable trimming module that estimates the dependency between instance-level representations and the global feature bank. Features highly correlated with global prototypes are softly suppressed, while instance-specific evidence is emphasized. This learnable mechanism encourages the model to prioritize causal signals over spurious correlations adaptively.
Experiments on VQA-RAD, SLAKE, SLAKE-CP and PathVQA demonstrate that LCT consistently improves robustness and generalization over existing debiasing strategies.
\keywords{Medical VQA  \and Causal Inference \and Causal Trimming \and Dynamic Feature Bank.}

\end{abstract}
\section{Introduction}

Medical Visual Question Answering (MedVQA) connects medical imaging with natural language to support clinical decision-making~\cite{cai2024counterfactual,liu2025cimb,zhan2023debiasing,tmilu}. Despite recent progress, existing models often suffer significant performance drops under Out-of-Distribution (OOD) settings. This issue largely arises from reliance on spurious correlations (e.g., associating question types with background textures) instead of true causal mechanisms~\cite{decoct}.

Recent approaches incorporate causal inference by modeling spurious correlations as confounders and applying backdoor adjustment with static dictionaries~\cite{11089950,liu2025cimb,zc}. However, spurious correlations arising from variable anatomical features are inherently dynamic, rendering static dictionaries inadequate. Meanwhile, Test-Time Adaptation (TTA) methods attempt to prune biased features during inference~\cite{liu2025testtime}, but they incur additional computational cost and treat the trained model as a black box, without improving intrinsic robustness.
To address these limitations, we propose Learnable Causal Trimming (LCT), an end-to-end framework that integrates causal intervention directly into training. Instead of static confounder sets, we construct a Dynamic Anatomical Feature Bank (DAFB) updated via momentum to capture evolving anatomical priors and dataset biases. A Causal Trimming (CT) module operates as a soft gate, estimating the dependency between instance features and DAFB prototypes, and adaptively reweighting confounder-correlated components while preserving informative evidence. This learnable intervention encourages the model to disentangle causal signals from confounding noise during optimization.

To our knowledge, we are the first to embed dynamic causal trimming as a trainable module within MedVQA. Our contributions are threefold:
\begin{itemize}

\item We propose the LCT framework, transforming causal adjustment in MedVQA from static or test-time strategies into an end-to-end trainable paradigm.
\item We design a Dynamic Anatomical Feature Bank with a Causal Trimming module to track and mitigate confounders during training without external annotations.
\item Experiments on four datasets demonstrate that LCT achieves state-of-the-art performance and improves OOD generalization.
\end{itemize}

\section{Methodology}

\subsection{Causal Formulation: The Confounding Problem}
\label{subsec:causal_formulation}

\begin{figure}[t]
\centering
\includegraphics[width=\textwidth]{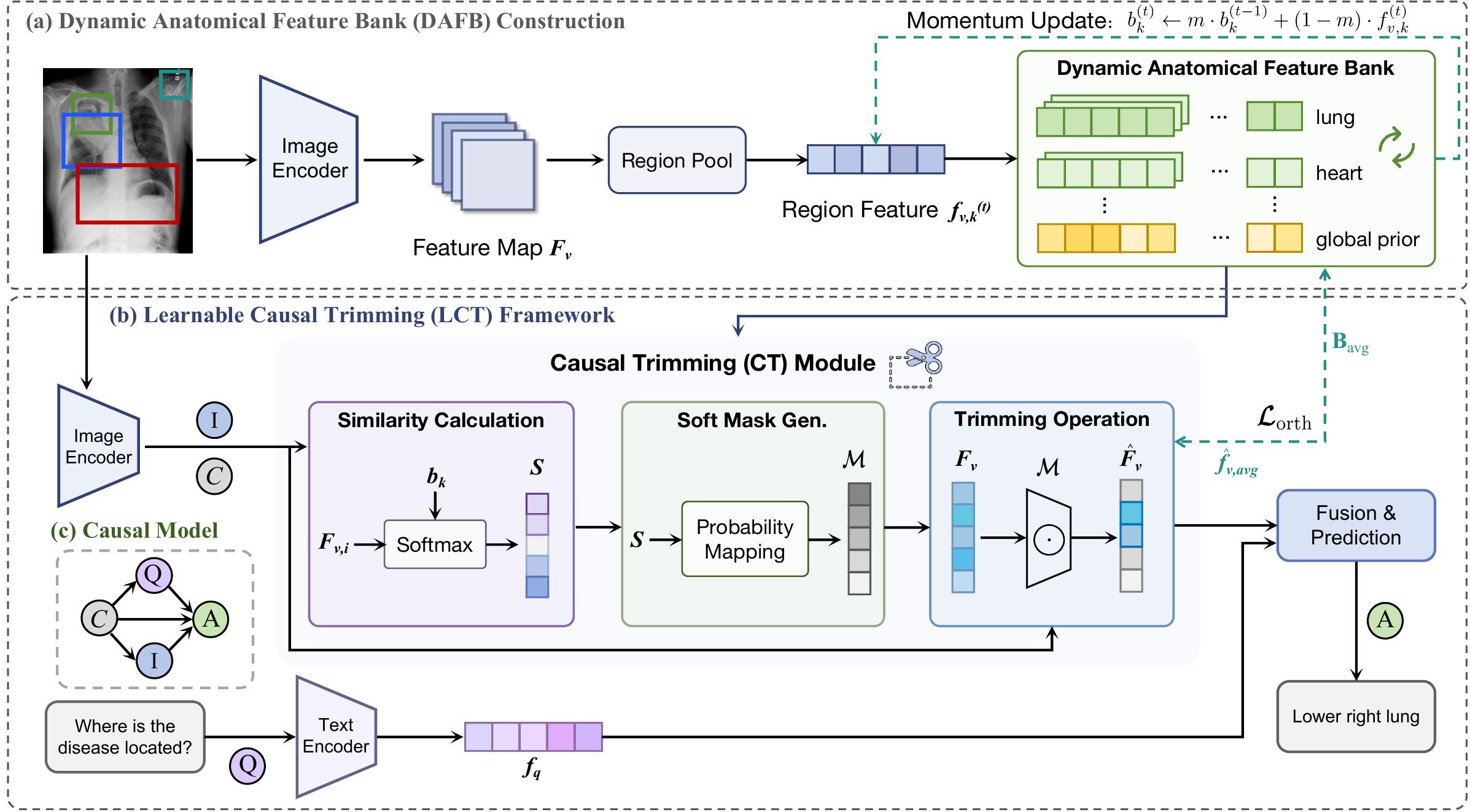}
\caption{
The proposed Learnable Causal Trimming (LCT) Framework with Dynamic Anatomical Feature Bank.
} \label{fig1}
\end{figure}

We analyze bias in MedVQA using a Structural Causal Model (SCM)~\cite{pearl2009causal}, as shown in Fig.~\ref{fig1}(c). $I$, $Q$, and $A$ denote the medical image, question, and answer. Ideally, the model should estimate $P(A|I, Q)$ through the causal paths $I\rightarrow A$ and $Q\rightarrow A$~\cite{tmi}. However, medical datasets contain strong priors. We introduce a latent confounder $C$ to represent dataset-level regularities. The backdoor paths $I \leftarrow C \rightarrow A$ and $Q \leftarrow C \rightarrow A$ induce spurious correlations. For example, if $C$ encodes lung-related priors, the model may predict ``Pneumonia'' from lung appearance alone, ignoring lesion-specific evidence in $I$. Standard supervised learning on the observational distribution may therefore capture shortcuts induced by $C$. To mitigate this issue, we simulate a causal intervention to reduce the confounding effect of $C$ on answer prediction. As $C$ is unobserved, we approximate it with a \textit{Dynamic Anatomical Feature Bank} and apply \textit{Causal Trimming} to reweight features associated with $C$ during representation learning.

\subsection{Learnable Causal Trimming (LCT) Framework}
\label{subsec:framework_overview}

Based on Sec.~\ref{subsec:causal_formulation}, the key to achieving robust MedVQA is executing the causal intervention by eliminating the confounding effect of $C$. To this end, we propose the Learnable Causal Trimming (LCT) framework. Unlike previous Test-Time Adaptation methods~\cite{liu2025testtime} that perform hard pruning on frozen models during inference, LCT internalizes the causal intervention mechanism into an end-to-end training pipeline. The overall architecture is depicted in Fig.~\ref{fig1}.

The framework processes the inputs through three main stages:

\textbf{Dual-Stream Encoding:} The input medical image $I$ and question $Q$ are processed by an image encoder and a text encoder to extract the regional visual feature map $F_v \in \mathbb{R}^{N \times D}$ and the question embedding $f_q$, respectively.

\textbf{Confounder Proxy Construction:} A confounder bank is maintained via momentum updates to aggregate global, high-frequency anatomical priors across the dataset, serving as a proxy of the unobserved confounder $C$.

\textbf{Causal Intervention via Trimming:} The Causal Trimming (CT) Module acts as a soft gate. It evaluates the similarity between the instance-specific visual features $F_v$ and the confounder prototypes in the DAFB, dynamically fading out spurious correlations to output the purified visual representation $\hat{F}_v$.

Finally, the trimmed visual features are fused with $f_q$ for answer prediction, supervised by a joint loss function that enforces causal disentanglement.

\subsection{Dynamic Anatomical Feature Bank (DAFB)}
\label{subsec:dafb}

As defined in our SCM, the confounder $C$ represents dataset-specific priors, such as the frequent co-occurrence of specific anatomical backgrounds and diagnostic answers. Since $C$ is latent and constantly evolving during training, we propose to explicitly model it using the \textbf{DAFB}.

Let $\mathbf{B} \in \mathbb{R}^{K \times D}$ denote the feature bank, where $K$ is the bank size and $D$ is the feature dimension. The bank $\mathbf{B} = \{b_1, b_2, \dots, b_K\}$ stores global prototypes representing common visual patterns (i.e., potential confounders). Instead of using a static dictionary, we employ a \textbf{Momentum Update} mechanism to maintain the stability of $\mathbf{B}$, ensuring it captures the long-term distribution of the dataset rather than the noisy fluctuations of a single mini-batch. Specifically, at training iteration $t$, we update the specific prototypes in the bank using the region-level features from the current batch. Let $f_{v,k}^{(t)}$ denote the aggregated visual feature of the current batch associated with the $k$-th prototype. We update the individual prototype $b_k \in \mathbf{B}$ via a momentum mechanism:
\begin{equation}
b_{k}^{(t)} \leftarrow m \cdot b_{k}^{(t-1)} + (1-m) \cdot f_{v,k}^{(t)},
\end{equation}
where $m \in [0,1)$ is a momentum coefficient. A large $m$ (e.g., 0.99) ensures that the bank evolves smoothly, effectively filtering out instance-specific details and retaining only the robust, high-frequency anatomical priors (e.g., generic lung shapes or background artifacts) that constitute the confounder $C$.

\subsection{Causal Trimming (CT) Module}
\label{subsec:trimming_module}

With the confounder proxy $\mathbf{B}$ constructed, the next critical step is to reduce the confounding effect of $C$ on the final prediction. To achieve this, we design the \textbf{Causal Trimming (CT) Module}, which adaptively reweights regional features according to their correlations with the dataset priors $\mathbf{B}$. As shown in Fig.~\ref{fig1}, this module operates in three sequential steps:

\textbf{1. Similarity Calculation:} Given the input visual feature map $F_v \in \mathbb{R}^{N \times D}$ (where $N$ corresponds to the number of patch-level feature tokens extracted by the visual encoder), we first compute the Confounder Relevance Score matrix $\mathbf{S} \in \mathbb{R}^{N \times K}$ by measuring the similarity between each image region and the bank prototypes:
\begin{equation}
    \mathbf{S}_{i,k} = \frac{\exp(F_{v,i} \cdot b_k^T / \tau)}{\sum_{j=1}^{K} \exp(F_{v,i} \cdot b_j^T / \tau)},
\end{equation}
where $\tau$ is a temperature hyperparameter. A high score $\mathbf{S}_{i,k}$ indicates that the $i$-th region strongly resembles a common prototype in the bank, implying it is likely associated with dataset-level priors.

\textbf{2. Soft Mask Generation:} To make the trimming process differentiable, we map the relevance scores into a Soft Trimming Mask $\mathcal{M} \in \mathbb{R}^{N \times 1}$. We aggregate the scores across all prototypes to determine the total confounding degree of each region:
\begin{equation}
    \mathcal{M}_i = 1 - \sigma \left( \psi \left( \sum_{k=1}^{K} \mathbf{S}_{i,k} \cdot w_k \right) \right),
    \label{eq:trimming_mask}
\end{equation}
where $\sigma(\cdot)$ is the Sigmoid function, $\psi$ is a learnable projection layer, and $w_k$ are learnable weights for the prototypes. The resulting mask $\mathcal{M}_i$ controls the residual gain of each region: it approaches 0 for regions highly correlated with confounder prototypes, reducing their additional amplification, and approaches 1 for unique, instance-specific diagnostic regions, enhancing their contribution.

\textbf{3. Trimming Operation:} Finally, the reweighted Trimmed Feature $\hat{F}_v$ is obtained via element-wise modulation combined with a residual connection:
\begin{equation}
    \hat{F}_v = F_v \odot \mathcal{M} + F_v.
    \label{eq:residual_trim}
\end{equation}
The residual connection is crucial here; it preserves gradient flow and retains the fundamental anatomical context, while the mask selectively emphasizes diagnostic evidence over confounder-correlated patterns.

\subsection{Optimization Objectives}
\label{subsec:loss}
To ensure accurate answers and causally independent representations, LCT optimizes a joint objective comprising a BCE classification loss ($\mathcal{L}_{\text{vqa}}$) and an orthogonality loss ($\mathcal{L}_{\text{orth}}$):
\begin{equation}
    \mathcal{L}_{\text{vqa}} = - \sum_{j=1}^{|\mathcal{A}|} y_j \log(P(a_j | \hat{F}_v, f_q)), \quad
    \mathcal{L}_{\text{orth}} = \left\| \frac{\hat{f}_{v, \text{avg}} \cdot \mathbf{B}_{\text{avg}}^T}{\|\hat{f}_{v, \text{avg}}\| \|\mathbf{B}_{\text{avg}}\|} \right\|^2,
\end{equation}
where $y$ is the ground truth. To explicitly enforce causal intervention, $\mathcal{L}_{\text{orth}}$ minimizes the cosine similarity between the mean vectors of the trimmed features ($\hat{f}_{v, \text{avg}}$) and the confounder bank ($\mathbf{B}_{\text{avg}}$), penalizing any residual correlation. The total objective is $\mathcal{L}_{\text{total}} = \mathcal{L}_{\text{vqa}} + \lambda \mathcal{L}_{\text{orth}}$, where $\lambda$ balances accuracy and causal disentanglement.

\section{Experiments}
\label{sec:experiments}

\subsection{Datasets and Implementation Details}
\label{subsec:datasets}

\textbf{Datasets.} \textbf{VQA-RAD} \cite{vqarad} contains 3,148 QA pairs on 315 radiology images. \textbf{SLAKE} \cite{liu2021slake} is a semantically-labeled bilingual dataset with 7,033 QA pairs on 642 radiology images. \textbf{PathVQA} \cite{he2020pathvqa} comprises 32,799 QA pairs on 4,998 pathology images, evaluating the model's generalizability across drastically different medical domains. Furthermore, to explicitly evaluate out-of-distribution (OOD) robustness, we utilize \textbf{SLAKE-CP} \cite{zhan2023debiasing}.

\textbf{Implementation Details.} Our framework is implemented in PyTorch. We leverage the pre-trained PMC-CLIP \cite{lin2023pmc} for dual-stream encoding to extract initial text embeddings and patch-level visual features. For the causal intervention, the DAFB is initialized with $K=256$ prototypes and updated with a momentum coefficient of $m$=0.99 to capture stable dataset priors. For the trimming module, the temperature scaling factor $\tau$ is set to 0.07. The entire network is trained end-to-end using the AdamW optimizer with a learning rate of $1e-4$ for 50 epochs. The balancing weight for the orthogonality loss, $\lambda$, is empirically set to 0.1.

\subsection{Comparison with State-of-the-Arts}
\label{subsec:sota}

Table \ref{tab:iid_sota_rad_slake} compares LCT with SOTA methods on VQA-RAD and SLAKE. LCT achieves the best Overall accuracy on both datasets, surpassing CIMB-MVQA by 1.8\% and 0.9\%, respectively.
For \textbf{Closed-ended questions}, LCT obtains the highest accuracy on VQA-RAD (87.9\%). Although M2I2 achieves 91.1\% on SLAKE Closed, it drops notably on Open questions (74.7\%). In contrast, LCT maintains balanced performance, achieving 89.9\% on Closed and a leading 83.4\% on Open questions.
For \textbf{Open questions}, LCT reaches 70.9\% on VQA-RAD and achieves SOTA on SLAKE Open. While slightly below MISS on VQA-RAD Open, LCT exceeds MISS by 5.1\% in Overall accuracy, indicating stronger holistic reasoning performance.

To further assess cross-modality generalization, we evaluate on PathVQA (Table \ref{tab:iid_sota_pathvqa}). LCT achieves the highest Overall accuracy of 65.6\%, outperforming MUMC (65.1\%). Notably, LCT establishes a new SOTA on Open-ended questions (41.1\%), while achieving comparable Closed accuracy (89.9\%). These results suggest that explicitly modeling dynamic confounders via causal trimming improves robustness, particularly for fine-grained pathological reasoning.

\newcommand{\std}[1]{{\scriptsize\textcolor{gray}{$\pm$ #1}}}

\begin{table*}[t]
\centering
\caption{Comparison of accuracy (\%) on VQA-RAD and SLAKE datasets. Best results are highlighted in \textbf{bold}.}
\label{tab:iid_sota_rad_slake}
\resizebox{0.8\textwidth}{!}{ 
\begin{tabular}{l ccc ccc}
\toprule
\multirow{2}{*}{\textbf{Methods}} & \multicolumn{3}{c}{\textbf{VQA-RAD}} & \multicolumn{3}{c}{\textbf{SLAKE}} \\
\cmidrule(lr){2-4} \cmidrule(lr){5-7}
 & Open & Closed & Overall & Open & Closed & Overall \\
\midrule
MEVF+SAN \cite{nguyen2019overcoming}     & 49.2 & 73.9 & 64.1 & 75.3 & 78.4 & 76.5 \\
MEVF+BAN \cite{nguyen2019overcoming}     & 49.2 & 77.2 & 66.1 & 77.8 & 79.8 & 78.6 \\
CPRD+BAN \cite{liu2021contrastive}     & 52.5 & 77.9 & 67.8 & 79.5 & 83.4 & 81.1 \\
M2I2 \cite{li2023self}         & 61.8 & 81.6 & 73.7 & 74.7 & \textbf{91.1} & 81.2 \\
MISS \cite{chen2024miss}         & \textbf{71.8} & 80.4 & 76.1 & 81.5 & 82.9 & 82.0 \\
M3AE \cite{chen2022m3ae}         & 67.2 & 83.5 & 77.0 & 80.3 & 87.8 & 83.3 \\
CCIS-MVQA \cite{cai2024counterfactual}    & 68.8 & 79.2 & 75.1 & 80.1 & 86.7 & 84.1 \\
PMC-CLIP \cite{lin2023pmc} & 67.0 & 84.0 & 77.6 & 81.9 & 88.0 & 84.3 \\
CLIPQCR \cite{eslami2023pubmedclip}      & 58.0\std{1.4} & 79.6\std{1.1} & 71.1\std{1.2} & 78.2\std{1.3} & 82.6\std{1.5} & 80.1\std{1.3} \\
DeBCF \cite{zhan2023debiasing}        & 58.6\std{1.1} & 80.9\std{0.8} & 71.6\std{1.0} & 80.8\std{0.9} & 84.9\std{0.7} & 82.6\std{0.9} \\
DeCoCT \cite{decoct} & 67.1\std{0.5} & 85.7\std{0.4} & 78.3\std{0.5} & 82.5\std{0.3} & 87.0\std{0.6} & 84.9\std{0.5} \\
CIMB-MVQA \cite{liu2025cimb} & 69.3\std{0.2} & 86.2\std{0.2} & 79.4\std{0.2} & 82.1\std{0.1} & 89.4\std{0.1} & 85.1\std{0.2} \\

\midrule
\textbf{LCT (Ours)} & 70.9\std{0.6} & \textbf{87.9}\std{0.4} & \textbf{81.2}\std{0.4} & \textbf{83.4}\std{0.3} & 89.9\std{0.5} & \textbf{86.0}\std{0.4} \\
\bottomrule
\end{tabular}
 }
\end{table*}

\begin{table*}[t]
\centering
\renewcommand{\arraystretch}{0.9}

\begin{minipage}[t]{0.42\textwidth} 
\centering
\caption{Comparison with SOTA methods on PathVQA.} 
\label{tab:iid_sota_pathvqa}
\resizebox{\textwidth}{!}{ 
\begin{tabular}{l ccc}
\toprule
\multirow{2}{*}{\textbf{Methods}} & \multicolumn{3}{c}{\textbf{PathVQA}} \\
\cmidrule(lr){2-4}
 & Open & Closed & Overall \\
\midrule
MEVF-BAN \cite{nguyen2019overcoming}          & 8.1 & 81.4 & 44.8 \\
MedGemma \cite{sellergren2025medgemma}  & 20.4 & 65.1 & 47.7 \\  
AMAM \cite{pan2022amam}           & 18.2 & 84.4 & 50.4 \\
LLaVA-Med \cite{li2023llava}         & 32.7 & 89.8 & 61.3 \\
M2I2 \cite{li2023self}     & 36.3  & 88.0   & 62.2 \\
CLIP-ViT \cite{van2023open}          & 40.0 & 87.0 & 63.6 \\
MUMC \cite{li2023masked}         & 39.0 & \textbf{90.4} & 65.1 \\
\midrule
\textbf{LCT (Ours)} & \textbf{41.1} & 89.9 & \textbf{65.6} \\
\bottomrule
\end{tabular}
}
\end{minipage}
\hfill 
\begin{minipage}[t]{0.52\textwidth} 
\centering
\caption{Comparison with SOTA methods on SLAKE-CP.}
\label{tab:ood_sota_comparison}
\resizebox{\textwidth}{!}{
\begin{tabular}{l ccc}
\toprule
\multirow{2}{*}{\textbf{Methods}} & \multicolumn{3}{c}{\textbf{SLAKE-CP}} \\
\cmidrule(lr){2-4}
 & Open & Closed & Overall \\
\midrule
MEVF+BAN \cite{nguyen2019overcoming}     & 13.0$_{\pm1.4}$ & 29.8$_{\pm1.2}$ & 29.1$_{\pm1.3}$ \\
CLIPQCR \cite{eslami2023pubmedclip}      & 13.4$_{\pm1.2}$ & 30.5$_{\pm1.1}$ & 30.0$_{\pm1.2}$ \\
CPRD+BAN \cite{liu2021contrastive}     & 13.9$_{\pm1.3}$ & 31.2$_{\pm1.5}$ & 30.4$_{\pm1.5}$ \\
DeBCF \cite{zhan2023debiasing}        & 18.6$_{\pm1.1}$ & 35.4$_{\pm1.0}$ & 34.2$_{\pm1.2}$ \\
MISS \cite{chen2024miss}      & 17.3$_{\pm1.1}$ & 49.2$_{\pm1.0}$ & 33.8$_{\pm1.1}$ \\
M2I2 \cite{li2023self}      & 17.3$_{\pm1.1}$ & 51.9$_{\pm0.9}$ & 35.2$_{\pm1.0}$ \\
M3AE \cite{chen2022m3ae}         & 24.4$_{\pm1.0}$ & 65.1$_{\pm0.9}$ & 45.4$_{\pm1.0}$ \\
DeCoCT \cite{decoct}       & 27.3$_{\pm0.5}$ & 69.6$_{\pm0.5}$ & 49.2$_{\pm0.6}$ \\
\midrule
\textbf{LCT (Ours)} & \textbf{28.9}$_{\pm0.6}$ & \textbf{71.6}$_{\pm0.4}$ & \textbf{51.0}$_{\pm0.5}$ \\
\bottomrule
\end{tabular}
}
\end{minipage}
\end{table*}

\subsection{Out-of-Distribution Generalization on SLAKE-CP}
Table \ref{tab:ood_sota_comparison} presents results on SLAKE-CP, a challenging OOD benchmark for evaluating diagnostic robustness. LCT achieves the best performance across all metrics, reaching 51.0\% Overall, 28.9\% on Open questions, and 71.6\% on Closed questions. These results indicate that the proposed method improves the model's ability to mitigate dataset-specific biases under distribution shifts.

Under the OOD setting, conventional representation learning models exhibit substantial performance degradation. For example, MISS and M2I2 drop to 33.8\% and 35.2\%, respectively. In comparison, LCT outperforms M2I2 by 15.8\% in Overall accuracy, demonstrating stronger robustness to distribution shifts.
Compared with recent debiasing and causal approaches, LCT also shows consistent improvements. Relative to DeCoCT, LCT improves Overall accuracy by 1.8\%, Open accuracy by 1.6\%, and Closed accuracy by 2.0\%. In addition, LCT maintains a low standard deviation (±0.5\% Overall), suggesting stable performance across runs. These findings support that the proposed learnable trimming strategy enhances both robustness and training stability in OOD scenarios.

\subsection{Ablation Study on Key Components}
\label{subsec:ablation}

We conduct ablation studies on VQA-RAD, SLAKE, and PathVQA to evaluate the individual contributions of each component. Results are in Table \ref{tab:ablation}.

\textbf{Baseline + DAFB (w/o CT):} 
We remove the causal trimming mechanism and directly concatenate DAFB features with the joint vision-language embeddings, followed by an MLP classifier. Using DAFB alone brings moderate improvements (+1.5\% on VQA-RAD and +1.3\% on PathVQA), suggesting that enriched anatomical priors are beneficial but insufficient to fully mitigate dataset biases when applied through standard feature fusion.

\textbf{Baseline + CT (w/o DAFB):} 
We remove the dynamic feature bank and compute trimming masks using global visual pooled features. Applying CT alone leads to larger gains(+3.5\% on VQA-RAD and +2.2\% on PathVQA). This indicates that explicitly suppressing spurious correlations contributes significantly to robustness, even without precise anatomical prototypes.

\textbf{Full LCT:}
Combining DAFB and CT achieves the best performance across all datasets. The consistent improvements over individual variants demonstrate that the two modules are complementary: DAFB enhances anatomical representation, while CT performs causal trimming. Their integration yields stronger and more stable diagnostic reasoning performance.

\begin{table*}[t]
\centering
\caption{Ablation study on the key components of the proposed LCT framework. \textbf{CT}: Causal Trimming module. \textbf{DAFB}: Dynamic Anatomical Feature Bank.}
\label{tab:ablation}

\resizebox{0.8\textwidth}{!}{
\begin{tabular}{ccc|ccc|ccc|c}
\toprule
\multirow{2}{*}{\textbf{Baseline}} & \multirow{2}{*}{\textbf{DAFB}} & \multirow{2}{*}{\textbf{CT}} & \multicolumn{3}{c|}{\textbf{VQA-RAD}} & \multicolumn{3}{c|}{\textbf{SLAKE}} & \textbf{PathVQA} \\
\cmidrule(lr){4-6} \cmidrule(lr){7-9} \cmidrule(lr){10-10}
 & & & Open & Closed & Overall & Open & Closed & Overall & Overall \\
\midrule
\checkmark & & & 66.5 & 83.1 & 76.5 & 81.6 & 87.7 & 84.0 & 61.4 \\
\checkmark & \checkmark & & 68.2 & 84.6 & 78.0 & 82.2 & 88.2 & 84.5 & 62.7 \\
\checkmark & & \checkmark & 69.3 & 87.1 & 80.0 & 82.2 & 88.9 & 84.8 & 63.6 \\
\checkmark & \checkmark & \checkmark & \textbf{70.9} & \textbf{87.9} & \textbf{81.2} & \textbf{83.4} & \textbf{89.9} & \textbf{86.0} & \textbf{65.6} \\
\bottomrule
\end{tabular}
}
\end{table*}

\subsection{Qualitative Analysis and Visualization}
\label{subsec:qualitative}

\begin{figure}[t]
 \centering
 \includegraphics[width=\textwidth]{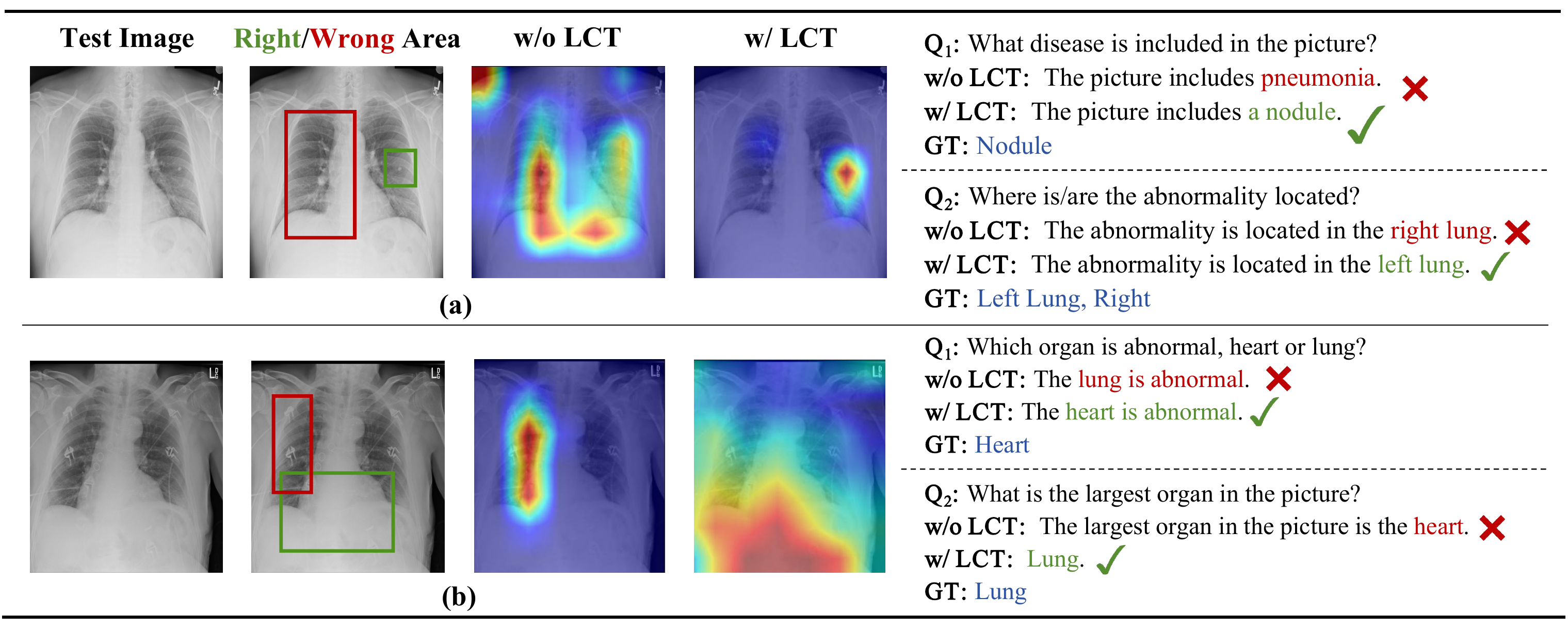}
 \caption{Qualitative comparison of Grad-CAM visualizations and QA results.}
 \label{fig:qualitative}
 \end{figure}

To illustrate how LCT mitigates spurious correlations, we present Grad-CAM visualizations and QA outputs in Figure~\ref{fig:qualitative}. Without LCT, the baseline often attends to irrelevant or biased regions. In (a), attention is diffusely distributed over the right lung (red box), failing to localize the actual lesion. In Example (b), the baseline focuses on an upper lung region instead of the target organs.

With LCT, the attention maps become more anatomically focused. In the ``w/ LCT'' column, activations concentrate on the relevant structures, such as the small nodule in the left lung (Example a) and the enlarged cardiac silhouette (Example b), indicating improved visual grounding.

The refined attention is reflected in the predicted answers. In Example (a), the baseline predicts ``pneumonia'' and incorrectly identifies the ``right lung'', whereas LCT correctly recognizes the ``nodule'' and localizes it in the ``left lung''. In Example (b), LCT accurately identifies the heart as the abnormal organ and the lung as the largest organ, while the baseline produces inconsistent responses. These results suggest that causal trimming improves anatomical grounding and reduces reliance on dataset-specific shortcuts.

\section{Conclusion}
\label{sec:conclusion}
In this paper, we introduced the Learnable Causal Trimming (LCT) framework to overcome the spurious correlations and dataset biases in MedVQA. By explicitly purifying visual representations at the feature level, LCT integrates a Dynamic Anatomical Feature Bank (DAFB) to continuously capture domain-specific medical prototypes and a Causal Trimming (CT) module to dynamically filter out confounding visual noise while preserving authentic pathological signals. Extensive experiments demonstrate that our approach not only establishes new state-of-the-art performance across multiple in-distribution benchmarks—including VQA-RAD, SLAKE, and PathVQA—but also exhibits exceptional robustness on the out-of-distribution SLAKE-CP dataset. Ultimately, LCT forces the network to ground its predictions in genuine anatomical evidence rather than exploiting statistical shortcuts, representing a significant step toward developing highly reliable multi-modal AI systems for clinical applications.

\begin{credits}
\subsubsection{\ackname} This work was supported in part by the National Natural Science
Foundation of China under Grant 62272337.

\subsubsection{\discintname} The authors have no competing interests to declare that
are relevant to the content of this article.

\end{credits}

\bibliographystyle{splncs04}
\bibliography{paper-0307}

\end{document}